\ifdefined\XeTeXversion\else
  \pdfoutput=1
\fi
\documentclass[11pt]{article}

\usepackage[final]{acl}

\usepackage{times}
\usepackage{latexsym}
\usepackage[T1]{fontenc}
\usepackage[utf8]{inputenc}
\usepackage{microtype}
\usepackage{inconsolata}
\usepackage{graphicx}

\usepackage{booktabs}
\usepackage{multirow}
\usepackage{tabularx}
\usepackage{array}
\usepackage{xcolor}
\usepackage{colortbl}
\usepackage{amsmath}

\newcommand{\benchmark}{CLBench-V}
\newcommand{\levelzero}{L0}
\newcommand{\levelone}{L1}
\newcommand{\leveltwo}{L2}

\title{\benchmark{}: Evaluating Multimodal Context Learning from Grounding to Knowledge Acquisition}

\author{%
\begin{tabular}{c}
  \begin{tabular}{c@{\hspace{2em}}c@{\hspace{2em}}c@{\hspace{2em}}c}
    \textbf{Lai Wei}\(^{1,2,*}\) &
    \textbf{Chengqi Li}\(^{1,3,*}\) &
    \textbf{Jiapeng Li}\(^{1,3,*}\) &
    \textbf{Ruina Hu}\(^{2,*}\)
  \end{tabular}\\[4pt]
  \begin{tabular}{c@{\hspace{3em}}c}
    \textbf{Yue Wang}\(^{2}\) &
    \textbf{Weiran Huang}\(^{1,3,\dagger}\)
  \end{tabular}\\[6pt]
{\normalfont \(^{1}\)School of Computer Science, Shanghai Jiao Tong University}\\
{\normalfont \(^{2}\)Zhongguancun Academy}\\
{\normalfont \(^{3}\)Shanghai Innovation Institute}
\end{tabular}%
}

\begin{document}
	\maketitle
\begingroup
\renewcommand{\thefootnote}{\fnsymbol{footnote}}
\footnotetext[1]{Equal contribution.\quad \(\dagger\) Correspondence to: Weiran Huang (\texttt{weiran.huang@sjtu.edu.cn}).}
\endgroup
	
	\begin{abstract}
		Real-world tasks often require models to learn from task-specific context rather than relying only on pre-trained knowledge.
		While recent work has highlighted this capability as context learning, existing evaluations mainly focus on textual contexts.
		In many practical settings, however, the context to be learned from is multimodal: scientific findings are conveyed through figures and tables, financial indicators are scattered across converted reports, and spatial decisions depend on maps, scenes, or web pages.
		We introduce \benchmark{}, a benchmark for multimodal context learning that addresses the difficulty of localizing where context use breaks down by organizing tasks around three dimensions: context grounding, new information application, and new knowledge learning.
		\benchmark{} combines converted public benchmarks with newly constructed datasets spanning domains such as science, finance, long-document understanding, spatial reasoning, and web-based visual question answering.
		To reduce the cost of constructing domain-specific context-learning tasks, we further use automated construction and filtering procedures for our newly built datasets.
		Across 3,443 instances and six recent multimodal models, the best overall score is only 0.2847, indicating that multimodal context learning remains far from saturated.
		Moreover, InternVL3.5-30B-A3B performs best on context grounding and new knowledge learning, while Qwen3.5-Plus performs best on new information application.
		We further analyze judge reliability, context length, image count, and representative failure cases. Code is available at this url.\footnote{\url{https://github.com/IamLihua/CLBench-V}}
	\end{abstract}
	
	\section{Introduction}
	
	\begin{figure*}[t]
		\centering
		\includegraphics[width=0.97\textwidth]{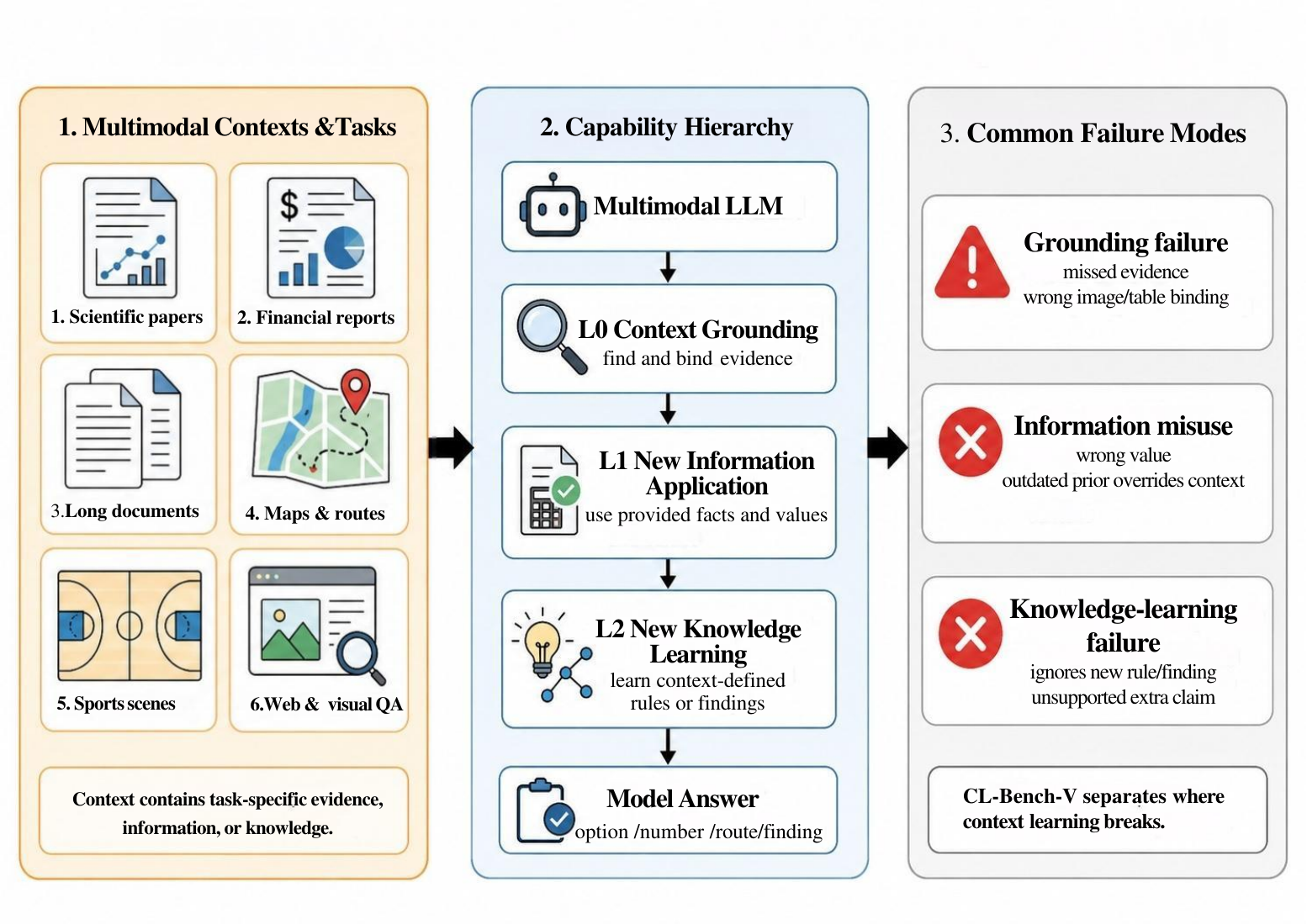}
		\caption{
			Overview of \benchmark{}.
			The benchmark covers diverse multimodal contexts and tasks, including scientific papers, financial reports, long documents, maps, sports scenes, and web-style visual question answering.
			We organize these tasks into a three-level capability hierarchy: context grounding, new information application, and new knowledge learning.
			The hierarchy helps separate common failures in multimodal context learning, such as missed evidence, information misuse, and ignoring context-defined knowledge.
		}
		\label{fig:overview}
	\end{figure*}
	
	Large language models increasingly solve problems using information provided at inference time rather than relying solely on parametric knowledge. A prominent training-free paradigm is \emph{context learning}, where models extract and apply task-specific facts, rules, procedures, or patterns from a given context. However, even when presented with identical contexts, models vary significantly in their ability to identify relevant information, apply it correctly, and avoid overriding it with prior knowledge. Long-context studies further show that evidence use can degrade with input length, evidence position, and the number of visual items \citep{liu-etal-2024-lost-middle,song-etal-2024-milebench,wang-etal-2024-mmniah}. Because real-world deployments frequently involve unseen documents, evolving reports, and dynamic environments, robust context learning capabilities are critical. Therefore, establishing a rigorous benchmark to evaluate how effectively models learn from and utilize provided context is essential.

	Existing context-learning benchmarks have driven significant progress \citep{bai-etal-2024-longbench,bai-etal-2025-longbench-v2,dou-etal-2026-clbench}. They evaluate whether models can follow new rules, process unfamiliar information, or avoid relying on prior knowledge. However, these benchmarks are restricted to unimodal textual scenarios, whereas many real-world contexts are inherently multimodal. Document and chart benchmarks test reading text, layout, and structured visual evidence \citep{mathew-etal-2021-docvqa,masry-etal-2022-chartqa}, while broader and multi-image suites test cross-domain reasoning and evidence binding \citep{yue-etal-2024-mmmu,liu-etal-2024-mmbench,wang-etal-2024-muirbench,meng-etal-2024-mmiu}. In domains such as science, finance, spatial reasoning, and web-based information seeking, task contexts frequently combine text with figures, tables, charts, and layouts \citep{wang-etal-2025-mmlongbench,selch-etal-2025-prismm,zhang-etal-2026-browsecomp-v3}. In these settings, models must first extract relevant evidence from diverse visual formats before applying it to reasoning. To address this gap, this paper focuses on establishing a rigorous benchmark for multimodal context learning.
	
	Extending context-learning evaluation to multimodal settings raises two challenges.
	First, prior context-learning benchmarks reveal recurring failures such as ignoring the provided context or misusing it, but they often make it difficult to localize which aspect of context use breaks down \citep{dou-etal-2026-clbench}. Controlled multimodal retrieval and memory probes expose similarly sharp failures under longer contexts and denser visual distractors \citep{wang-etal-2024-mmneedle,ren2026memlens,guo2026memeye}.
	Motivated by these observations, we extend the failure analysis to multimodal contexts, where errors can arise from failing to ground relevant evidence, misusing newly provided information, or ignoring knowledge defined only in the context.
	We therefore organize \benchmark{} around three dimensions of multimodal context learning, making it naturally suited for analyzing where model performance degrades.
	Second, high-quality context-learning tasks are expensive to annotate manually. Existing multi-image instruction data demonstrate the value of systematic task construction, but they do not by themselves guarantee long-context dependence or diagnostic evaluation \citep{jiang-etal-2024-mantis}. For our newly constructed domain-specific tasks, we therefore introduce automated data construction and filtering procedures that synthesize candidate instances at scale while retaining task-specific quality control.
	
	We propose \benchmark{}, a multimodal context-learning benchmark organized around a three-level hierarchy.
	\levelzero{} \emph{Context Grounding} tests whether models can locate, read, and combine evidence from multimodal contexts.
	\levelone{} \emph{New Information Application} tests whether models can apply newly provided facts, values, or cases using capabilities they already possess.
	\leveltwo{} \emph{New Knowledge Learning} tests whether models can acquire new rules, discoveries, task semantics, or empirical conclusions from the context itself.
	This hierarchy is designed to separate three common failure modes: failing to access the relevant multimodal context, misusing context that has been accessed, and ignoring newly defined knowledge in favor of prior knowledge.
	To instantiate the hierarchy, we combine converted public benchmarks with newly constructed domain-specific tasks, including recent medical-paper conclusion inference and financial-report ROE analysis, and evaluate all tasks under a unified inference and scoring framework.
	Figure~\ref{fig:overview} summarizes this design.
	
	Experiments on six multimodal models across 3,443 instances show that current systems still struggle with multimodal context learning, with the best overall score reaching only 0.2847.
	InternVL3.5-30B-A3B performs best on \levelzero{} Context Grounding and \leveltwo{} New Knowledge Learning, while Qwen3.5-Plus performs best on \levelone{} New Information Application.
	This level-wise variation suggests that accepting multimodal input is not sufficient for reliable context use; models may ground visual evidence, apply newly supplied information, and acquire context-defined knowledge with very different levels of reliability.
	We also compare different LLM judges and analyze performance across context length, image count, answer types, and representative failure modes.
	
	Our contributions are:
	\begin{itemize}
		\item We define multimodal context learning as a hierarchical capability covering context grounding, new information application, and new knowledge learning.
		\item We construct \benchmark{}, a multi-source benchmark that combines converted public benchmarks with newly built tasks from recent medical papers and financial reports.
		\item We evaluate a diverse set of multimodal models and analyze judge reliability, context length, image count, answer types, and common failure modes.
	\end{itemize}
	
	\section{Related Work}
	
	\paragraph{Context learning and long-context evaluation.}
	Recent long-context benchmarks test retrieval, long-document question answering, summarization, few-shot learning, and realistic long-context reasoning across documents, code, and structured data \citep{bai-etal-2024-longbench,bai-etal-2025-longbench-v2}.
	Position-sensitive analyses show that nominal context capacity can overstate effective evidence use, especially when relevant information appears in the middle of a long input \citep{liu-etal-2024-lost-middle}.
	Synthetic tests such as needle-in-a-haystack and its extensions further probe whether models can retrieve and manipulate information buried in long distractor contexts \citep{hsieh-etal-2024-ruler}.
	Another line of work studies many-shot in-context learning, where expanded context windows allow models to learn from hundreds or thousands of demonstrations at inference time \citep{agarwal-etal-2024-many-shot}.
	CL-Bench further argues that models should be evaluated on whether they can acquire new information, rules, procedures, and empirical patterns from context, beyond retrieval or simple demonstration learning \citep{dou-etal-2026-clbench}.
	\benchmark{} extends this motivation to multimodal settings, where the relevant context may be distributed across text, images, tables, figures, and document-derived layouts.
	
	\paragraph{Multimodal long-context modeling.}
	Recent multimodal long-context work studies how training recipes, sequence-length exposure, and data-mixture design affect transfer beyond nominal context windows \citep{wang-etal-2025-mmlongbench,li-etal-2025-giraffe}.
	Evaluation has expanded from realistic multi-image contexts in MileBench to page-rich document settings in MMLongBench-Doc and M-LongDoc \citep{song-etal-2024-milebench,ma-etal-2024-mmlongbench-doc,chia-etal-2024-mlongdoc}.
	Controlled visual needle tests isolate retrieval, counting, evidence-position, and distractor sensitivity, while newer memory suites measure retention over extended multimodal streams \citep{wang-etal-2024-mmniah,wang-etal-2024-mmneedle,ren2026memlens,guo2026memeye}.
	These studies reveal that context capacity and context learning are not equivalent: models may access long inputs yet still fail to internalize context-defined rules, resolve cross-modal conflicts, or transfer newly provided knowledge to final answers.
	Our benchmark is complementary to this line by directly testing whether long-context-capable multimodal models can \emph{learn from} context under strict success criteria.

	\paragraph{Multi-image and visually rich document understanding.}
	Multi-image benchmarks test comparison, co-reference, temporal ordering, and cross-image composition, and instruction-tuning resources provide supervision for interleaved visual contexts \citep{wang-etal-2024-muirbench,meng-etal-2024-mmiu,jiang-etal-2024-mantis}.
	For documents, DocVQA and ChartQA establish core reading and visual-numerical reasoning settings, while broad multimodal suites such as MMMU and MMBench cover diverse expert and general-purpose tasks \citep{mathew-etal-2021-docvqa,masry-etal-2022-chartqa,yue-etal-2024-mmmu,liu-etal-2024-mmbench}.
	Retrieval-oriented systems and benchmarks further emphasize page selection and visual-layout preservation in long documents \citep{dong-etal-2025-mmdocir,shi-etal-2024-visrag,faysse-etal-2024-colpali}.
	In contrast, \benchmark{} uses these perceptual and retrieval demands as prerequisites, then diagnoses whether grounded evidence is merely found, applied as new information, or generalized as newly acquired knowledge.

	\section{\benchmark{}: Task and Benchmark Design}
	\label{sec:benchmark}
	
	\subsection{Overview}
	
	\benchmark{} is designed to evaluate models' ability to learn from provided multimodal context and apply what they learn to solve tasks in realistic scenarios.
	The required knowledge is largely beyond what models can rely on from pre-training alone, and appears in diverse forms, including rendered document pages, Markdown-converted long documents and financial reports, scientific figures and tables, charts, maps, sports-scene images, web-style screenshots, and multi-image evidence.
	All necessary information is organized in the provided context, so models are evaluated on context learning rather than external retrieval.
	
	To make this ability measurable and diagnostic, \benchmark{} organizes tasks into a three-level hierarchy:
	\levelzero{} Context Grounding, \levelone{} New Information Application, and \leveltwo{} New Knowledge Learning.
	This design isolates whether failures come from missing evidence, misusing context-specific information, or failing to acquire context-defined knowledge.
	It also highlights the hardest regime, \leveltwo{}, where models must synthesize and transfer newly learned knowledge from multimodal context.
	
	Finally, all datasets are evaluated in a unified conversion, inference, and scoring framework with task-specific metrics.
	This setup supports cross-level comparability while preserving fine-grained diagnosis of failure modes.
	The next subsection details the hierarchy, followed by benchmark construction and evaluation protocol.
	
	\subsection{Three-Level Capability Hierarchy}
	
	Building on the overview above, we organize multimodal context learning with a three-level hierarchy.
	The hierarchy is defined by what the supplied context contributes to the task: evidence that must be grounded, new information that must be applied, or new knowledge that must be learned.
	Figure~\ref{fig:overview} gives a compact visualization of this progression.
	
	\paragraph{\levelzero{}: Context Grounding.}
	At this level, the model must locate, read, and combine evidence from newly provided multimodal context, without needing to learn a new rule system or domain theory.
	Accordingly, these tasks primarily test evidence access and grounding fidelity, including visual search, counting, spatial localization, topology following, and cross-image binding.
	Concretely, this level includes transit-map reasoning from ReasonMap, high-resolution visual search from Insight-O3, hard visual QA from ZeroBench, sports spatial intelligence from CourtSI, and multi-image scientific figure reasoning from MIRBench.
	
	\paragraph{\levelone{}: New Information Application.}
	At this level, the model must use new facts, values, cases, or document-specific information from the context while relying on reasoning skills it already has.
	In practice, these tasks test whether the model can correctly bind context-specific values and use them in existing reasoning procedures, while avoiding outdated or irrelevant prior knowledge.
	This level includes visual fact QA from Pix2Fact, long-document QA from MMLongBench-Pic, browsing-style final-answer tasks from BrowseComp-V3, scientific inconsistency questions from PRISMM-Bench, and financial report ROE analysis.
	
	\paragraph{\leveltwo{}: New Knowledge Learning.}
	At this level, the model must acquire new rules, empirical patterns, task semantics, scientific conclusions, or procedures from the context and apply them to answer the task.
	Compared with lower levels, these tasks require knowledge acquisition from context-defined evidence rather than only evidence retrieval or value application.
	This level includes the image-rendered table subset converted from CL-Bench and paper conclusion inference.
	Compared with \levelzero{} and \levelone{}, \leveltwo{} is the most challenging because the model must synthesize context-defined knowledge and transfer it to answer generation, not just locate evidence or apply known formulas.
	
	The levels are cumulative: failures in grounding can prevent information use, and failures in information use can prevent knowledge learning.
	However, the levels are analytically distinct and expose different failure modes.
	In addition to assigning each task to a level, we analyze instances along orthogonal dimensions such as model-side requirements, multimodal operations, context complexity, answer format, and evaluation type.
	These dimensions make the benchmark diagnostic: two tasks may share the same level while stressing different operations, such as chart reading, route topology validation, page-level search, conclusion induction, or numeric evaluation.
	
	With this hierarchy and diagnostic view in place, we next describe how \benchmark{} is constructed.
	
	\subsection{Benchmark Construction}
	
	\benchmark{} is built from two complementary sources: integrated public benchmarks and newly constructed domain-specific tasks.
	Rather than treating source datasets as isolated silos, we map both sources into one normalized schema and organize tasks by the capability they primarily diagnose: context grounding, new information application, or new knowledge learning.
	Table~\ref{tab:datasets} summarizes the current composition under this unified organization.
	For evaluation, we keep deterministic task-specific evaluators for structured outputs and use judge-based semantic evaluation only for open-ended outputs.
	We describe these two construction pipelines in order: \textit{Data Integration} followed by \textit{Data Construction}.

\begin{table*}[t]
	\centering
	\caption{Current composition of \benchmark{} organized by the proposed capability levels. Some datasets are marked as boundary cases when their instances combine grounding with higher-level information use or knowledge learning.}
	\label{tab:datasets}
	\begingroup
	\small
	\renewcommand{\arraystretch}{1.10}
	\setlength{\tabcolsep}{5pt}
	\begin{tabularx}{\textwidth}{c l r | X l}
		\toprule[1.2pt]
		\multicolumn{3}{c|}{\textbf{Benchmark Component}} & \multicolumn{2}{c}{\textbf{Task Specification}} \\
		\cmidrule(lr){1-3} \cmidrule(lr){4-5}
		\textbf{Level} & \textbf{Dataset} & \textbf{N} & \textbf{Task Form} & \textbf{Evaluation} \\
		\midrule[0.8pt]
		\levelzero{} & ReasonMap-Short & 103 & Subway map route planning with short queries & Topology judge \\
		\levelzero{} & ReasonMap-Long & 101 & Subway map route planning with long queries & Topology judge \\
		\levelzero{} & ReasonMap-Plus & 37 & Subway map counting, boolean, and MCQ tasks & Rule-based extraction \\
		\levelzero{} & Insight-O3 & 145 & High-resolution visual search and multi-hop MCQ & Option extraction \\
		\levelzero{} & ZeroBench & 43 & Hard visual question answering & Normalized / numeric match \\
		\levelzero{} & CourtSI & 645 & Sports spatial intelligence & MA / MRA / LocA \\
		\levelzero{}--\levelone{} & MIRBench & 1063 & Multi-image scientific figure reasoning MCQ & Option extraction \\
		\midrule
		\levelone{} & Pix2Fact & 43 & Multimodal visual fact QA & LLM judge / normalized match \\
		\levelone{} & MMLongBench-Pic & 78 & Long-context multimodal QA with page images & LLM judge / baseline match \\
		\levelone{} & BrowseComp-V3 & 57 & Multimodal browsing final-answer tasks & LLM judge / normalized match \\
		\levelone{} & Financial Report ROE & 206 & Multimodal financial report analysis & ROE exact / MAE \\
		\levelone{}--\leveltwo{} & PRISMM-Bench & 245 & Scientific paper inconsistency QA & Option extraction \\
		\midrule
		\leveltwo{} & CL-Bench-Table & 96 & Context learning with image-rendered tables & Rubric judge \\
		\leveltwo{} & Paper Conclusion & 581 & Scientific paper conclusion inference & List-level semantic judge \\
		\midrule[0.8pt]
		\rowcolor{blue!5} -- & \textbf{Total} & \textbf{3,443} & -- & -- \\
		\bottomrule[1.2pt]
	\end{tabularx}
	\endgroup
\end{table*}

	\subsubsection{Data Integration}
	
	For the public-benchmark source, we first convert candidate datasets into a unified input-output format, then apply explicit curation criteria before final admission.
	
	\textbf{Criteria for unified conversion.}
	These criteria follow the same objective as our capability design: evaluating context learning rather than shortcut retrieval.
	Concretely, we apply \textbf{four} criteria.
	First, \emph{visual dependency}: images must provide task-critical evidence, so the target answer should not be reliably solvable by text-only OCR snippets, generic image captions, or other image-free shortcuts.
	Second, \emph{low irrelevant context}: we minimize unrelated context content whenever possible, so evaluation does not collapse into pure long-context needle-search and remains focused on new information application and new knowledge learning. This distinction is important because controlled needle benchmarks already isolate retrieval under evidence-position and distractor-density changes \citep{wang-etal-2024-mmniah,wang-etal-2024-mmneedle}.
	Third, \emph{multi-hop preference}: we prioritize samples that require compositional reasoning across multiple evidence pieces (e.g., cross-region, cross-image, cross-modal, or cross-field binding), reducing single-span lookup shortcuts. This follows the broader shift from isolated image QA toward cross-image and expert-level compositional evaluation \citep{wang-etal-2024-muirbench,yue-etal-2024-mmmu}.
	Fourth, \emph{scope exclusion}: we do not include video tasks or quasi-video tasks with dense contiguous multi-frame reasoning. These tasks are often highly domain-specific, supported by very large-scale data and mature dedicated evaluation ecosystems, and many recent models have already been exposed to video data during training, giving them baseline competence in this regime. In contrast, our focus is multimodal context-learning ability that goes beyond what models can solve by relying primarily on training alone.
	
	\textbf{Data filter.}
	We first choose source benchmarks that collectively cover visual fact verification, long-context document understanding, scientific inconsistency reasoning, transit-map planning, sports spatial intelligence, high-resolution visual search, multi-image reasoning, hard visual QA, browsing-style final-answer synthesis, and context-defined rule following, including Pix2Fact, MMLongBench, PRISMM-Bench, ReasonMap, InSight-o3, MIRBench, ZeroBench, CourtSI-Bench, BrowseComp-$V^3$, and CL-Bench \citep{jiang-etal-2026-pix2fact,wang-etal-2025-mmlongbench,selch-etal-2025-prismm,feng-etal-2025-reasonmap,li-etal-2025-insight-o3,du-etal-2025-mir,roberts-etal-2025-zerobench,yang-etal-2026-courtsi,zhang-etal-2026-browsecomp-v3,dou-etal-2026-clbench}.
	We then run conversion-time rejection sampling with Qwen3.5-Plus as the primary inspector and remove candidate instances that are likely solvable through shortcut paths (e.g., weak visual dependency, excessive irrelevant context, or single-span lookup behavior).
	After this automatic filtering stage, we conduct manual review to retain high-quality samples that satisfy our four criteria, preserve diversity of operations and answer types, and remain aligned with the benchmark focus on grounding, new information application, and new knowledge acquisition.
	For retained tasks, we finalize normalization by embedding options for MCQ items, preserving both Markdown and image-rendered document variants when available, keeping topology fields for route-structure validation, and moving required search evidence into context when the original benchmark assumes external retrieval.
	This pipeline produces the integrated public-benchmark portion of \benchmark{}.
	
	\subsubsection{Data Construction}
	
	Beyond public-benchmark integration, we construct domain-specific tasks that target capability gaps under-covered by existing sources, while keeping the same schema and evaluator interface.
	This source contributes two tasks: financial report ROE analysis and paper conclusion inference.
	
	\paragraph{Financial report ROE analysis.}
	We collect recent financial reports and convert each PDF into a Markdown-style multimodal document representation.
	Models are asked to perform DuPont-style reasoning and compute ROE from the converted report context.
	Prior financial QA benchmarks establish numerical reasoning over text, tables, and conversational evidence \citep{chen-etal-2021-finqa,zhu-etal-2021-tatqa,chen-etal-2022-convfinqa}; our setting extends this line to long, converted reports in which relevant quantities can be distributed across pages and visual structures.
	This task primarily evaluates new information application: the financial formulas are standard, but the relevant values are document-specific.
	The task is built around the standard three-factor DuPont decomposition:
	\begin{equation}
		\begin{aligned}
			\mathrm{ROE}
			&=
			\mathrm{Net\ Profit\ Margin}
			\times \mathrm{Asset\ Turnover} \\
			&\quad \times
			\mathrm{Equity\ Multiplier}.
		\end{aligned}
		\label{eq:dupont}
	\end{equation}
	Equivalently, the model must identify and combine quantities such as net income, revenue, total assets, and shareholders' equity from the converted report content.
	Although the current evaluation focuses on final ROE exact match, the task requires document-level grounding, entity-value binding, and numerical reasoning over financial statements.
	
	\paragraph{Paper conclusion inference.}
	We collect recent medical papers and construct tasks that ask models to infer paper-specific scientific findings.
	Research-paper QA datasets such as QASPER ground information-seeking questions in scientific articles \citep{dasigi-etal-2021-qasper}, while recent multimodal benchmarks additionally stress scientific figures, cross-image evidence, and internal paper consistency \citep{du-etal-2025-mir,selch-etal-2025-prismm}.
	We focus on recent papers to reduce the chance that their results have appeared in model pretraining data, thereby mitigating information leakage.
	We choose the medical domain because biomedical papers often place dense empirical evidence in figures, microscopy images, radiology images, plots, and tables, making them a strong testbed for extracting phenomena and conclusions from multimodal evidence after PDF-to-Markdown conversion.
	
	For each paper, we first convert the PDF into a Markdown-style multimodal representation that preserves extracted text and figure/table references, and then create two truncated versions.
	The reference-construction version includes the paper up to and including the Result section, and is used to generate a set of reference scientific findings.
	The evaluation version includes only the content before the Result section, while retaining the available preceding text and figures.
	At test time, a model receives the evaluation version and is asked to predict the scientific findings that would be supported by the paper's results.
	This setup requires the model to synthesize experimental motivation, methods, extracted figures/tables, and result-related multimodal evidence, rather than simply copying explicit conclusions.
	
	We filter low-quality instances and retain 581 clean examples.
	For scoring, let $N_{\mathrm{ref}}$ be the number of reference findings, $N_{\mathrm{hit}}$ be the number of model-predicted findings judged correct, and $N_{\mathrm{extra}}$ be the number of unsupported extra findings.
	The paper-level score is
	\begin{equation}
		s_{\mathrm{paper}} =
		\max\left(0, \frac{N_{\mathrm{hit}} - N_{\mathrm{extra}}}{N_{\mathrm{ref}}}\right).
		\label{eq:paper-score}
	\end{equation}
	The subtraction term penalizes over-prediction: a model should not receive high credit by listing many speculative findings in the hope that some match the reference set.
	
	Together, the two pipelines provide complementary coverage: the integration source contributes broad capability diversity, and the construction source contributes targeted high-value settings for context-grounded information use and knowledge acquisition.
	We next describe the unified evaluation protocol used across all tasks.
	
	\subsection{Evaluation Protocol}
	
	All inference and evaluation logic is implemented within a unified framework.
	For every dataset, the framework constructs standardized model inputs, records the same metadata fields, invokes models through a common interface, extracts the final answer when needed, and then dispatches the prediction to the corresponding evaluator.
	This design keeps prompting, logging, answer extraction, and score aggregation consistent across datasets, while still allowing task-specific metrics for different answer formats.
	
	\benchmark{} uses task-specific evaluation whenever possible.
	Multiple-choice datasets use option extraction followed by exact matching.
	Numeric tasks use normalized exact match, tolerance-based matching, or mean absolute error.
	Route-planning tasks use a topology judge that verifies endpoints, valid lines, and transfer continuity.
	Open-ended tasks use answer extraction followed by rubric-based or semantic LLM judging.
	For paper conclusion inference, we use a list-level judge to estimate $N_{\mathrm{hit}}$ and $N_{\mathrm{extra}}$, then compute the paper-level score in Equation~\ref{eq:paper-score}.
	
	This design reflects a conservative principle carried throughout this section: deterministic evaluators are preferred for structured outputs, while LLM judges are reserved for tasks where semantic equivalence is necessary.
	
	\section{Experiments}
	\label{sec:experiments}
	
	\subsection{Experimental Setup}
	\label{sec:setup}
	
	\paragraph{Models.}
	
	We currently report preliminary results for six multimodal models: InternVL3.5-30B-A3B, Qwen3.6-27B, Qwen3.5-Plus, Doubao-Seed-2.0-Lite, Kimi-K2.6, and GPT-5.4.
	Qwen3.6-27B is also used as the main judge for judge-based tasks.
	Additional model runs are ongoing.
	All models receive the same task prompt, input context, and answer-format instruction unless a model-specific API constraint requires minor formatting changes.
	
	\paragraph{Prompting and answer extraction.}
	
	Prompts are designed to be task-faithful and to minimize ambiguity in the expected answer format.
	For multiple-choice tasks, models are instructed to provide a final option letter.
	For open-ended tasks, models are instructed to provide a concise final answer after reasoning.
	For judge-based tasks, we extract the final answer before judging to reduce sensitivity to irrelevant explanation text.
	
	\subsection{Main Results}
	\label{sec:main-results}
	
	\paragraph{Results by level.}
	
	Table~\ref{tab:main-level} reports model performance aggregated by the three levels.
	For judge-based tasks, the main results use Qwen3.6-27B as the judge model.
	Overall and level scores are sample-weighted over the currently completed subsets.

\begin{table*}[t]
	\centering
	\caption{Main results aggregated by the proposed multimodal context learning levels. Dataset scores are weighted by their corresponding numbers of instances. For aggregation, MIRBench is assigned to \levelzero{} and PRISMM-Bench to \levelone{}. The best result in each column is shown in \textbf{bold}, and the second-best is \underline{underlined}.}
	\label{tab:main-level}
	\resizebox{0.78\textwidth}{!}{%
	\renewcommand{\arraystretch}{1.10}
	\setlength{\tabcolsep}{8pt}
	\begin{tabular}{l | c | c c c}
		\toprule[1.2pt]
		\multirow{2}{*}{\textbf{Model}} & \cellcolor{blue!5}\textbf{Aggregate} & \multicolumn{3}{c}{\textbf{Capability Level}} \\
		\cmidrule(lr){3-5}
		& \cellcolor{blue!5}\textbf{Overall} & \textbf{\levelzero{} Grounding} & \textbf{\levelone{} Information} & \textbf{\leveltwo{} Knowledge} \\
		\midrule[0.8pt]
		InternVL3.5-30B-A3B & \cellcolor{blue!5}\textbf{0.2847} & \textbf{0.3080} & 0.1313 & \textbf{0.3536} \\
		Qwen3.6-27B & \cellcolor{blue!5}0.1958 & 0.1865 & \underline{0.2878} & 0.1398 \\
		Qwen3.5-Plus & \cellcolor{blue!5}0.1862 & 0.1507 & \textbf{0.2954} & 0.1964 \\
		Doubao-Seed-2.0-Lite & \cellcolor{blue!5}0.1850 & 0.1797 & 0.2241 & 0.1655 \\
		Kimi-K2.6 & \cellcolor{blue!5}\underline{0.1991} & 0.1755 & 0.2582 & \underline{0.2186} \\
		GPT-5.4 & \cellcolor{blue!5}0.1894 & \underline{0.2003} & 0.2814 & 0.0694 \\
		\bottomrule[1.2pt]
	\end{tabular}
	}
\end{table*}

\begin{figure*}[t]
    \centering
    \makebox[\textwidth][c]{%
        \hspace*{-1cm}%
        \includegraphics[width=0.82\textwidth]
        {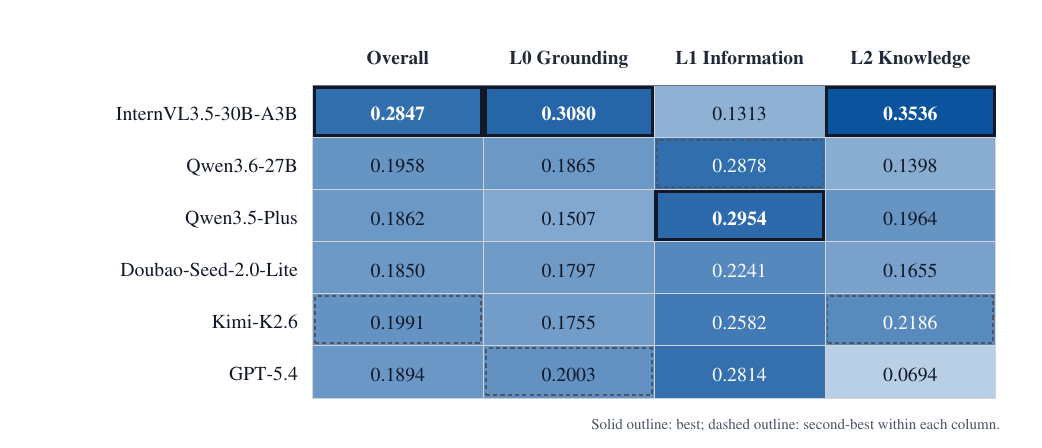}%
        \hspace*{1cm}%
    }

    \caption{Capability profiles of the evaluated models. Cell color encodes the score; solid outlines mark the best model in each column and dashed outlines mark the second-best.}
    \label{fig:capability-profile}
\end{figure*}

	The overall scores remain low for all evaluated models, with the best model reaching 0.2847.
	This suggests that multimodal context learning remains challenging even for recent multimodal systems, and that the benchmark is not saturated by any single model family.
	The level-wise breakdown further shows that models fail in different ways.
	InternVL3.5-30B-A3B obtains the highest overall score and the strongest \levelzero{} grounding score, while Qwen3.5-Plus achieves the strongest \levelone{} information score.
	InternVL3.5-30B-A3B also achieves the best \leveltwo{} score, largely due to its strong performance on paper conclusion inference, but struggles on \levelone{} tasks that require applying document-specific values and facts.
	Qwen3.6-27B, Qwen3.5-Plus, and Doubao-Seed-2.0-Lite are comparatively stronger on \levelone{}, indicating better use of newly supplied information than acquisition of context-defined knowledge.
	Figure~\ref{fig:capability-profile} makes these contrasting capability profiles visually explicit.
	
	\paragraph{Results by dataset.}
	
	Table~\ref{tab:dataset-results} reports dataset-level performance.
	This view helps separate broad cross-level trends from dataset-specific bottlenecks such as route topology, sports localization, long-document reading, or scientific conclusion induction.

\begin{table*}[t]
	\centering
	\caption{Dataset-level results for completed subsets, grouped by capability level. Boundary datasets are shown in separate columns. The best result in each column is shown in \textbf{bold}, and the second-best is \underline{underlined}. For InternVL3.5-30B-A3B, 0.0000 on Financial Report ROE indicates that the run exceeded the model's maximum context length.}
	\label{tab:dataset-results}
	\resizebox{\textwidth}{!}{%
	\renewcommand{\arraystretch}{1.10}
	\setlength{\tabcolsep}{3.5pt}
	\begin{tabular}{l | *{6}{c} | c | *{4}{c} | c | *{2}{c}}
		\toprule[1.2pt]
		\multirow{2}{*}{\textbf{Model}} & \multicolumn{6}{c|}{\textbf{\levelzero{} Grounding}} & \multicolumn{1}{c|}{\textbf{\levelzero{}--\levelone{}}} & \multicolumn{4}{c|}{\textbf{\levelone{} Information}} & \multicolumn{1}{c|}{\textbf{\levelone{}--\leveltwo{}}} & \multicolumn{2}{c}{\textbf{\leveltwo{} Knowledge}} \\
		\cmidrule(lr){2-7} \cmidrule(lr){8-8} \cmidrule(lr){9-12} \cmidrule(lr){13-13} \cmidrule(lr){14-15}
		& \textbf{O3} & \textbf{RMap-S} & \textbf{RMap-L} & \textbf{RMap+} & \textbf{ZB} & \textbf{CourtSI} & \textbf{MIR} & \textbf{Pix2Fact} & \textbf{MML-Pic} & \textbf{Browse} & \textbf{Finance} & \textbf{PRISMM} & \textbf{CL} & \textbf{Paper} \\
		\midrule[0.8pt]
		InternVL3.5-30B-A3B & 0.2483 & 0.0680 & 0.0792 & 0.2162 & 0.0465 & 0.1336 & \textbf{0.4807} & 0.0588 & 0.0260 & 0.0000 & 0.0000 & 0.3184 & 0.0000 & \textbf{0.4120} \\
		Qwen3.6-27B & \textbf{0.5724} & 0.2913 & 0.3168 & 0.4865 & \underline{0.1395} & 0.1281 & 0.1383 & \textbf{0.2941} & \underline{0.3506} & 0.0877 & 0.0922 & \underline{0.4776} & \textbf{0.0938} & 0.1474 \\
		Qwen3.5-Plus & \underline{0.4276} & 0.2427 & 0.2376 & \underline{0.5676} & 0.1282 & \underline{0.1389} & 0.0894 & 0.2000 & \textbf{0.3621} & 0.1404 & 0.0825 & \textbf{0.5061} & 0.0000 & 0.2289 \\
		Doubao-Seed-2.0-Lite & 0.4069 & 0.3495 & \underline{0.3861} & 0.1892 & \textbf{0.2093} & 0.1115 & 0.1524 & 0.1628 & 0.2676 & 0.1754 & 0.0588 & 0.3714 & 0.0000 & 0.1928 \\
		Kimi-K2.6 & 0.3931 & \textbf{0.6800} & \textbf{0.4059} & \textbf{0.5946} & 0.1026 & 0.1169 & 0.0990 & 0.0556 & 0.2051 & \underline{0.1800} & \underline{0.1165} & 0.4480 & 0.0000 & \underline{0.2547} \\
		GPT-5.4 & 0.3724 & \underline{0.3689} & 0.2772 & 0.4865 & \textbf{0.2093} & \textbf{0.1783} & \underline{0.1562} & \underline{0.2326} & 0.2564 & \textbf{0.2105} & \textbf{0.1262} & 0.4449 & \underline{0.0104} & 0.0792 \\
		\bottomrule[1.2pt]
	\end{tabular}
	}
\end{table*}

	The dataset-level results reveal substantial heterogeneity beneath the aggregate scores.
	No model dominates across all datasets: Qwen3.6-27B leads on Pix2Fact, Insight-O3, and CL-Bench-Table; Qwen3.5-Plus on MMLongBench-Pic and PRISMM-Bench; Kimi-K2.6 on all three ReasonMap subsets; InternVL3.5-30B-A3B on Paper Conclusion and MIRBench; and GPT-5.4 on BrowseComp-V3, CourtSI, and Financial Report ROE, while tying Doubao-Seed-2.0-Lite on ZeroBench.
	Several tasks remain difficult across models, especially Pix2Fact, BrowseComp-V3, CL-Bench-Table, and CourtSI, suggesting persistent weaknesses in visual fact grounding, web-style evidence synthesis, context-defined rule following, and spatial intelligence.
	The sharp variation across datasets supports the need for evaluating multimodal context learning as a structured set of capabilities rather than as a single averaged score.
	
	\section{Do LLM Judges Matter?}
	\label{sec:judge}
	
	Because \benchmark{} contains both structured and open-ended tasks, evaluation quality is itself an important experimental question.
	We measure score shifts across judge models on datasets that require semantic evaluation.
	The current comparison reports macro scores for two prediction sets.
	
	\begin{table}[t]
		\centering
		\caption{Comparison of LLM judges on semantically evaluated tasks. Scores are macro averages over datasets requiring LLM judging for each prediction set. The best result in each column is shown in \textbf{bold}, and the second-best is \underline{underlined}.}
		\label{tab:judge-comparison}
		\begingroup
		\small
		\renewcommand{\arraystretch}{1.10}
		\setlength{\tabcolsep}{3.5pt}
		\begin{tabularx}{\columnwidth}{X | c c | c}
			\toprule[1.2pt]
			\textbf{Judge} & \textbf{InternVL} & \textbf{Qwen3.6} & \cellcolor{blue!5}\textbf{Avg.} \\
			\midrule[0.8pt]
			Qwen3-VL-4B-Instruct & \textbf{0.1095} & \textbf{0.3148} & \cellcolor{blue!5}\textbf{0.2122} \\
			Qwen3-VL-8B-Instruct & 0.0671 & 0.2744 & \cellcolor{blue!5}0.1708 \\
			Qwen3-30B-A3B-Instruct-2507 & \underline{0.0742} & 0.2676 & \cellcolor{blue!5}0.1709 \\
			Qwen3-VL-32B-Instruct & 0.0705 & 0.2423 & \cellcolor{blue!5}0.1564 \\
			Qwen3.6-27B & 0.0686 & \underline{0.2867} & \cellcolor{blue!5}\underline{0.1777} \\
			\bottomrule[1.2pt]
		\end{tabularx}
		\endgroup
	\end{table}
	
	The comparison shows non-negligible score variation across judge models.
	Qwen3-VL-4B-Instruct gives the highest average score in this setting, while Qwen3-VL-32B-Instruct gives the lowest average score.
	This confirms that judge choice can affect reported performance on open-ended multimodal context-learning tasks, motivating transparent reporting of the judge model and further calibration in future versions.
	
	\section{Diagnostic Analysis}
	\label{sec:analysis}
	
	\subsection{Context Complexity}
	
	We analyze whether model performance is correlated with two simple measures of input complexity: the token length of the context and the number of images supplied in the context.
	For Qwen3.6-27B, token length is essentially uncorrelated with score (Pearson $r=-0.0042$, Spearman $\rho=0.0500$), and image count is weakly negative under both measures (Pearson $r=-0.0756$, Spearman $\rho=-0.0739$).
	For InternVL3.5-30B-A3B, both measures are negative under Pearson correlation ($r=-0.1220$ for token length and $r=-0.1652$ for image count) but weakly positive under Spearman correlation ($\rho=0.0280$ and $\rho=0.1217$).
	The disagreement in sign is itself informative: it indicates that the negative linear correlations are produced by a small tail of extremely long, image-dense inputs rather than by a steady decline in accuracy as inputs grow.
	Excluding the financial reports, which are by far the longest inputs and on which InternVL3.5-30B-A3B exceeds its context limit, the token-score Pearson correlation turns positive for both models ($r=0.0510$ for InternVL3.5-30B-A3B and $r=0.1429$ for Qwen3.6-27B).
	These results suggest that input length alone does not explain current failures: what matters is whether an input crosses a model's usable context limit, not how long it is within that limit.
	Image count remains model-dependent. In rank terms InternVL3.5-30B-A3B improves as more images are supplied ($\rho=0.1217$, rising to $\rho=0.2317$ once the financial reports are excluded), whereas Qwen3.6-27B does not ($\rho=-0.0739$ and $\rho=-0.0402$), suggesting that additional views act as richer evidence for one model and as visual distraction for the other.
	
	\subsection{Failure Taxonomy}
	
	We manually inspect representative bad cases and categorize failures into six types:
	\emph{evidence missing}, where the model does not attend to the relevant visual region;
	\emph{evidence misbinding}, where evidence is attached to the wrong entity, figure, page, or option;
	\emph{context misuse}, where the model reads the relevant information but applies it incorrectly;
	\emph{prior override}, where the model relies on parametric knowledge despite contradictory context;
	\emph{incomplete induction}, where the model notices local facts but fails to infer the broader context-defined rule or conclusion;
	and \emph{format failure}, where the answer violates the expected output schema.
	Detailed, representative case studies that operationalize these failure modes across L0/L1/L2 context-learning levels are provided in Appendix~\ref{sec:failure-case-studies}.

	\section{Limitations}
	
	\benchmark{} is an initial benchmark and has several limitations.
	First, it integrates heterogeneous data sources whose annotation styles and evaluation protocols differ.
	Second, some subsets are small and should be interpreted as diagnostic probes rather than comprehensive task distributions.
	Third, LLM judges are necessary for several open-ended tasks but may introduce model-specific biases despite answer extraction and rubric design.
	Fourth, our current financial-report task evaluates only final ROE values and does not yet fully score intermediate DuPont reasoning steps.
	
	\section{Conclusion}
	
	We introduced \benchmark{}, a benchmark for multimodal context learning.
	By organizing tasks into context grounding, new information application, and new knowledge learning, \benchmark{} distinguishes failures of visual access from failures of contextual reasoning and knowledge acquisition.
	The benchmark combines converted public datasets with newly constructed scientific-paper and financial-report tasks, and pairs them with task-specific evaluation protocols.
	Our planned and preliminary experiments analyze model performance, judge reliability, input complexity, and representative bad cases.
	We hope \benchmark{} provides a useful diagnostic framework for building multimodal systems that can not only see context, but also use and learn from it.
	
	\section*{Ethics Statement}
	
	\paragraph{Source Data.}
	\benchmark{} is built from converted public benchmarks and newly constructed tasks based on publicly available scientific papers and financial reports.
	For public benchmark components, we follow the intended research use of the original datasets and retain source attribution in our data records.
	For newly constructed scientific-paper tasks, we use recent publicly available papers and derive questions and reference findings from paper content.
	For financial-report tasks, we use publicly released corporate reports and evaluate only numerical reasoning over disclosed financial statements.
	Although some scientific papers may include medical images or biomedical findings, the benchmark is designed only for model evaluation and should not be used for clinical decision-making, diagnosis, investment advice, or other high-stakes deployment.
	
	\paragraph{Ethics Review.}
	Our benchmark does not involve human-subject experiments, private user data, or newly collected personal information.
	We nevertheless screen converted examples for obvious personally identifiable information and exclude content when it is not necessary for the task.
	The tasks are intended to evaluate whether models can use provided context; they are not intended to infer sensitive attributes about individuals or to support decisions about real people.
	We will release the data with documentation describing source provenance, task construction, evaluation scripts, and known limitations.
	
	\paragraph{AI Assistant Usage.}
	AI assistants were used to support data-processing and benchmark-construction programming, including format conversion, script drafting, prompt organization, and preliminary quality checks.
	All benchmark definitions, filtering criteria, evaluation protocols, and paper claims were reviewed by the authors.
	AI-generated code or text was treated as an aid rather than an authority, and outputs used in the benchmark were subject to manual inspection or task-specific validation before inclusion.

	\bibliography{custom}

\begin{thebibliography}{37}
\providecommand{\natexlab}[1]{#1}

\bibitem[{Agarwal et~al.(2024)Agarwal, Singh, Zhang, Bohnet, Rosias, Chan,
  Zhang, Anand, Abbas, Nova, Co-Reyes, Chu, Behbahani, Faust, and
  Larochelle}]{agarwal-etal-2024-many-shot}
Rishabh Agarwal, Avi Singh, Lei~M. Zhang, Bernd Bohnet, Luis Rosias, Stephanie
  Chan, Biao Zhang, Ankesh Anand, Zaheer Abbas, Azade Nova, John~D. Co-Reyes,
  Eric Chu, Feryal Behbahani, Aleksandra Faust, and Hugo Larochelle. 2024.
\newblock \href {https://arxiv.org/abs/2404.11018} {Many-shot in-context
  learning}.
\newblock \emph{arXiv preprint arXiv:2404.11018}.

\bibitem[{Bai et~al.(2024)Bai, Lv, Zhang, Lyu, Tang, Huang, Du, Liu, Zeng, Hou,
  Dong, Tang, and Li}]{bai-etal-2024-longbench}
Yushi Bai, Xin Lv, Jiajie Zhang, Hongchang Lyu, Jiankai Tang, Zhidian Huang,
  Zhengxiao Du, Xiao Liu, Aohan Zeng, Lei Hou, Yuxiao Dong, Jie Tang, and
  Juanzi Li. 2024.
\newblock \href {https://arxiv.org/abs/2308.14508} {{LongBench}: A bilingual,
  multitask benchmark for long context understanding}.
\newblock \emph{arXiv preprint arXiv:2308.14508}.

\bibitem[{Bai et~al.(2025)Bai, Tu, Zhang, Peng, Wang, Lv, Cao, Xu, Hou, Dong,
  Tang, and Li}]{bai-etal-2025-longbench-v2}
Yushi Bai, Shangqing Tu, Jiajie Zhang, Hao Peng, Xiaozhi Wang, Xin Lv, Shulin
  Cao, Jiazheng Xu, Lei Hou, Yuxiao Dong, Jie Tang, and Juanzi Li. 2025.
\newblock \href {https://arxiv.org/abs/2412.15204} {{LongBench v2}: Towards
  deeper understanding and reasoning on realistic long-context multitasks}.
\newblock \emph{arXiv preprint arXiv:2412.15204}.

\bibitem[{Chen et~al.(2021)Chen, Chen, Smiley, Shah, Borova, Langdon, Moussa,
  Beane, Huang, Routledge, and Wang}]{chen-etal-2021-finqa}
Zhiyu Chen, Wenhu Chen, Charese Smiley, Sameena Shah, Iana Borova, Dylan
  Langdon, Reema Moussa, Matt Beane, Ting-Hao Huang, Bryan Routledge, and
  William~Yang Wang. 2021.
\newblock \href {https://doi.org/10.18653/v1/2021.emnlp-main.300} {{FinQA}: A
  dataset of numerical reasoning over financial data}.
\newblock In \emph{Proceedings of the 2021 Conference on Empirical Methods in
  Natural Language Processing}.

\bibitem[{Chen et~al.(2022)Chen, Li, Smiley, Ma, Shah, and
  Wang}]{chen-etal-2022-convfinqa}
Zhiyu Chen, Shiyang Li, Charese Smiley, Zhiqiang Ma, Sameena Shah, and
  William~Yang Wang. 2022.
\newblock \href {https://doi.org/10.18653/v1/2022.emnlp-main.421} {{ConvFinQA}:
  Exploring the chain of numerical reasoning in conversational finance question
  answering}.
\newblock In \emph{Proceedings of the 2022 Conference on Empirical Methods in
  Natural Language Processing}.

\bibitem[{Chia et~al.(2024)Chia, Cheng, Chan, Liu, Song, Aljunied, Poria, and
  Bing}]{chia-etal-2024-mlongdoc}
Yew~Ken Chia, Liying Cheng, Hou~Pong Chan, Chaoqun Liu, Maojia Song,
  Sharifah~Mahani Aljunied, Soujanya Poria, and Lidong Bing. 2024.
\newblock \href {https://arxiv.org/abs/2411.06176} {{M-LongDoc}: A benchmark
  for multimodal super-long document understanding and a retrieval-aware tuning
  framework}.
\newblock \emph{arXiv preprint arXiv:2411.06176}.

\bibitem[{Dasigi et~al.(2021)Dasigi, Lo, Beltagy, Cohan, Smith, and
  Gardner}]{dasigi-etal-2021-qasper}
Pradeep Dasigi, Kyle Lo, Iz~Beltagy, Arman Cohan, Noah~A. Smith, and Matt
  Gardner. 2021.
\newblock \href {https://arxiv.org/abs/2105.03011} {A dataset of
  information-seeking questions and answers anchored in research papers}.
\newblock In \emph{Proceedings of the 2021 Conference of the North American
  Chapter of the Association for Computational Linguistics}.

\bibitem[{Dong et~al.(2025)Dong, Chang, Goh, Li, Tang, and
  Liu}]{dong-etal-2025-mmdocir}
Kuicai Dong, Yujing Chang, Xin~Deik Goh, Dexun Li, Ruiming Tang, and Yong Liu.
  2025.
\newblock \href {https://arxiv.org/abs/2501.08828} {{MMDocIR}: Benchmarking
  multimodal retrieval for long documents}.
\newblock \emph{arXiv preprint arXiv:2501.08828}.

\bibitem[{Dou et~al.(2026)Dou, Zhang, Yin, Huang, Shen, Wang, Chen, Ni, Ye,
  Zhang, Xie, Hu, Wang, Wang, Xiao, Liu, Xu, Guo, Zhou, Gui, Wu, Qiu, Zhang,
  Huang, Jiang, Wang, and Yao}]{dou-etal-2026-clbench}
Shihan Dou, Ming Zhang, Zhangyue Yin, Chenhao Huang, Yujiong Shen, Junzhe Wang,
  Jiayi Chen, Yuchen Ni, Junjie Ye, Cheng Zhang, Huaibing Xie, Jianglu Hu,
  Shaolei Wang, Weichao Wang, Yanling Xiao, Yiting Liu, Zenan Xu, Zhen Guo,
  Pluto Zhou, and 8 others. 2026.
\newblock \href {https://arxiv.org/abs/2602.03587} {{CL-bench}: A benchmark for
  context learning}.
\newblock \emph{arXiv preprint arXiv:2602.03587}.

\bibitem[{Du et~al.(2025)Du, Zhang, Nan, Deng, Chen, Zhang, Xiao, Huang, Pan,
  Qi, and Leng}]{du-etal-2025-mir}
Hang Du, Jiayang Zhang, Guoshun Nan, Wendi Deng, Zhenyan Chen, Chenyang Zhang,
  Wang Xiao, Shan Huang, Yuqi Pan, Tao Qi, and Sicong Leng. 2025.
\newblock \href {https://arxiv.org/abs/2509.17040} {From easy to hard: The
  {MIR} benchmark for progressive interleaved multi-image reasoning}.
\newblock \emph{arXiv preprint arXiv:2509.17040}.

\bibitem[{Faysse et~al.(2024)Faysse, Sibille, Wu, Omrani, Viaud, Hudelot, and
  Colombo}]{faysse-etal-2024-colpali}
Manuel Faysse, Hugues Sibille, Tony~F. Wu, Bilel Omrani, Gautier Viaud, Celine
  Hudelot, and Pierre Colombo. 2024.
\newblock \href {https://arxiv.org/abs/2407.01449} {{ColPali}: Efficient
  document retrieval with vision language models}.
\newblock \emph{arXiv preprint arXiv:2407.01449}.

\bibitem[{Feng et~al.(2025)Feng, Wang, Ouyang, Kong, Song, Zhu, Wang, and
  Wang}]{feng-etal-2025-reasonmap}
Sicheng Feng, Song Wang, Shuyi Ouyang, Lingdong Kong, Zikai Song, Jianke Zhu,
  Huan Wang, and Xinchao Wang. 2025.
\newblock \href {https://arxiv.org/abs/2505.18675} {{ReasonMap}: Towards
  fine-grained visual reasoning from transit maps}.
\newblock \emph{arXiv preprint arXiv:2505.18675}.

\bibitem[{Guo et~al.(2026)Guo, Wang, Yao, Zhang, Song, Wang, He, Wu, Lin, Wei,
  Liao, Liu, Liu, Lin, Liu, Zhou, Wu, Zhou, Zhao, Zhang, Liu, Lin, Chen, and
  Zhang}]{guo2026memeye}
Jiaheng Guo, Ruidong Wang, Chunlei Yao, Chenyang Zhang, Yaqi Song, Zichao Wang,
  Kun He, Shiguang Wu, Ziliang Lin, Qipeng Wei, Xuan Liao, Xihui Liu, Li~Liu,
  Ying-Ying Lin, Wenbing Liu, Shuo Zhou, Qiang Wu, Ping Zhou, Bingxuan Zhao,
  and 5 others. 2026.
\newblock \href {https://arxiv.org/abs/2605.10187} {{MemEye}: Benchmarking
  multimodal large language models with long-context visual memory}.
\newblock \emph{arXiv preprint arXiv:2605.10187}.

\bibitem[{Hsieh et~al.(2024)Hsieh, Sun, Kriman, Acharya, Rekesh, Jia, Zhang,
  and Ginsburg}]{hsieh-etal-2024-ruler}
Cheng-Ping Hsieh, Simeng Sun, Samuel Kriman, Shantanu Acharya, Dima Rekesh, Fei
  Jia, Yang Zhang, and Boris Ginsburg. 2024.
\newblock \href {https://arxiv.org/abs/2404.06654} {{RULER}: What's the real
  context size of your long-context language models?}
\newblock \emph{arXiv preprint arXiv:2404.06654}.

\bibitem[{Jiang et~al.(2024)Jiang, He, Zeng, Wei, Ku, Liu, and
  Chen}]{jiang-etal-2024-mantis}
Dongfu Jiang, Xuan He, Huaye Zeng, Cong Wei, Max Ku, Qian Liu, and Wenhu Chen.
  2024.
\newblock \href {https://arxiv.org/abs/2405.01483} {{MANTIS}: Interleaved
  multi-image instruction tuning}.
\newblock \emph{arXiv preprint arXiv:2405.01483}.

\bibitem[{Jiang et~al.(2026)Jiang, Zhang, Zhang, Zheng, Yang, Wang, and
  Ong}]{jiang-etal-2026-pix2fact}
Yifan Jiang, Cong Zhang, Bofei Zhang, Qiaofeng Zheng, Yifan Yang, Bingzhang
  Wang, and Yew-Soon Ong. 2026.
\newblock \href {https://arxiv.org/abs/2602.00593} {{Pix2Fact}: When vision is
  not enough -- benchmarking fine-grained {VQA} with web verification on
  high-resolution real-world scenes}.
\newblock \emph{arXiv preprint arXiv:2602.00593}.

\bibitem[{Li et~al.(2025{\natexlab{a}})Li, Yao, Wu, Yu, Chen, Bai, Hou, Hong,
  Zhang, and Zhang}]{li-etal-2025-insight-o3}
Kaican Li, Lewei Yao, Jiannan Wu, Tiezheng Yu, Jierun Chen, Haoli Bai, Lu~Hou,
  Lanqing Hong, Wei Zhang, and Nevin~L. Zhang. 2025{\natexlab{a}}.
\newblock \href {https://arxiv.org/abs/2512.18745} {{InSight-o3}: Empowering
  multimodal foundation models with generalized visual search}.
\newblock \emph{arXiv preprint arXiv:2512.18745}.

\bibitem[{Li et~al.(2025{\natexlab{b}})Li, Huang, Wang, Yang, Liu
  et~al.}]{li-etal-2025-giraffe}
Zhuowan Li, Dong Huang, Wenhao Wang, Shenglong Yang, Zhe Liu, and 1 others.
  2025{\natexlab{b}}.
\newblock \href {https://aclanthology.org/2025.acl-long.1351/} {{GIRAFFE}:
  Design choices for extending the context length of visual language models}.
\newblock In \emph{Proceedings of the 63rd Annual Meeting of the Association
  for Computational Linguistics}.

\bibitem[{Liu et~al.(2024{\natexlab{a}})Liu, Lin, Hewitt, Paranjape,
  Bevilacqua, Petroni, and Liang}]{liu-etal-2024-lost-middle}
Nelson~F. Liu, Kevin Lin, John Hewitt, Ashwin Paranjape, Michele Bevilacqua,
  Fabio Petroni, and Percy Liang. 2024{\natexlab{a}}.
\newblock \href {https://doi.org/10.1162/tacl_a_00638} {Lost in the middle: How
  language models use long contexts}.
\newblock \emph{Transactions of the Association for Computational Linguistics},
  12:157--173.

\bibitem[{Liu et~al.(2024{\natexlab{b}})Liu, Duan, Zhang, Li, Zhang, Zhao,
  Yuan, Wang, He, Liu, Chen, and Lin}]{liu-etal-2024-mmbench}
Yuan Liu, Haodong Duan, Yuanhan Zhang, Bo~Li, Songyang Zhang, Wangbo Zhao, Yike
  Yuan, Jiaqi Wang, Conghui He, Ziwei Liu, Kai Chen, and Dahua Lin.
  2024{\natexlab{b}}.
\newblock \href {https://arxiv.org/abs/2307.06281} {{MMBench}: Is your
  multi-modal model an all-around player?}
\newblock \emph{arXiv preprint arXiv:2307.06281}.

\bibitem[{Ma et~al.(2024)Ma, Zang, Chen, Chen, Jiao, Li, Lu
  et~al.}]{ma-etal-2024-mmlongbench-doc}
Yubo Ma, Yuhang Zang, Liangyu Chen, Meiqi Chen, Yizhu Jiao, Xinze Li, Xinyuan
  Lu, and 1 others. 2024.
\newblock \href {https://arxiv.org/abs/2407.01523} {{MMLongBench-Doc}:
  Benchmarking long-context document understanding with visualizations}.
\newblock \emph{arXiv preprint arXiv:2407.01523}.

\bibitem[{Masry et~al.(2022)Masry, Long, Tan, Joty, and
  Hoque}]{masry-etal-2022-chartqa}
Ahmed Masry, Do~Xuan Long, Jia~Qing Tan, Shafiq Joty, and Enamul Hoque. 2022.
\newblock \href {https://arxiv.org/abs/2203.10244} {{ChartQA}: A benchmark for
  question answering about charts with visual and logical reasoning}.
\newblock In \emph{Findings of the Association for Computational Linguistics:
  ACL 2022}.

\bibitem[{Mathew et~al.(2021)Mathew, Karatzas, and
  Jawahar}]{mathew-etal-2021-docvqa}
Minesh Mathew, Dimosthenis Karatzas, and C.~V. Jawahar. 2021.
\newblock \href {https://arxiv.org/abs/2007.00398} {{DocVQA}: A dataset for
  {VQA} on document images}.
\newblock In \emph{Proceedings of the IEEE/CVF Winter Conference on
  Applications of Computer Vision}.

\bibitem[{Meng et~al.(2024)Meng, Wang, Li, Lu, Tian, Liao, Zhu, Dai
  et~al.}]{meng-etal-2024-mmiu}
Fanqing Meng, Jin Wang, Chuanhao Li, Quanfeng Lu, Hao Tian, Jiaqi Liao, Xizhou
  Zhu, Jifeng Dai, and 1 others. 2024.
\newblock \href {https://arxiv.org/abs/2408.02718} {{MMIU}: Multimodal
  multi-image understanding for evaluating large vision-language models}.
\newblock \emph{arXiv preprint arXiv:2408.02718}.

\bibitem[{Ren et~al.(2026)Ren, An, Zhang, Zheng, Zhang, Yan, Zhang, Zhang, Liu,
  Wu, Chen, Yan, and Liu}]{ren2026memlens}
Yilong Ren, Jingjing An, Luming Zhang, Mingjie Zheng, Wei Zhang, Songlin Yan,
  Hongyu Zhang, Wenqi Zhang, Kai Liu, Di~Wu, Hao Chen, Wang Yan, and Yuanchun
  Liu. 2026.
\newblock \href {https://arxiv.org/abs/2605.14906} {{MemLens}: Evaluating
  memory in multimodal large language models}.
\newblock \emph{arXiv preprint arXiv:2605.14906}.

\bibitem[{Roberts et~al.(2025)Roberts, Taesiri, Sharma, Gupta, Roberts,
  Croitoru, Bogolin, Tang, Langer, Raina, Raina, Xiong, Udandarao, Lu, Chen,
  Purkis, Yan, Lin, Shin, Yang, Nguyen, Atkinson, Baranwal, Coca, Dang,
  Dziadzio, Kunz, Liang, Lo, Pulfer, Walton, Yang, Han, and
  Albanie}]{roberts-etal-2025-zerobench}
Jonathan Roberts, Mohammad~Reza Taesiri, Ansh Sharma, Akash Gupta, Samuel
  Roberts, Ioana Croitoru, Simion-Vlad Bogolin, Jialu Tang, Florian Langer,
  Vyas Raina, Vatsal Raina, Hanyi Xiong, Vishaal Udandarao, Jingyi Lu, Shiyang
  Chen, Sam Purkis, Tianshuo Yan, Wenye Lin, Gyungin Shin, and 15 others. 2025.
\newblock \href {https://arxiv.org/abs/2502.09696} {{ZeroBench}: An impossible
  visual benchmark for contemporary large multimodal models}.
\newblock \emph{arXiv preprint arXiv:2502.09696}.

\bibitem[{Selch et~al.(2025)Selch, Hou, Mirza, Doveh, Glass, Feris, and
  Lin}]{selch-etal-2025-prismm}
Lukas Selch, Yufang Hou, M.~Jehanzeb Mirza, Sivan Doveh, James Glass, Rogerio
  Feris, and Wei Lin. 2025.
\newblock \href {https://arxiv.org/abs/2510.16505} {{PRISMM-Bench}: A benchmark
  of peer-review grounded multimodal inconsistencies}.
\newblock \emph{arXiv preprint arXiv:2510.16505}.

\bibitem[{Shi et~al.(2024)Shi, Tang, Xu, Cui, Ran, Yan, Liu, Wang, Han, Liu,
  and Sun}]{shi-etal-2024-visrag}
Yu~Shi, Chaoyue Tang, Bokai Xu, Junbo Cui, Junhao Ran, Yukun Yan, Zhenghao Liu,
  Shuo Wang, Xu~Han, Zhiyuan Liu, and Maosong Sun. 2024.
\newblock \href {https://arxiv.org/abs/2410.10594} {{VisRAG}: Vision-based
  retrieval-augmented generation on multi-modality documents}.
\newblock \emph{arXiv preprint arXiv:2410.10594}.

\bibitem[{Song et~al.(2024)Song, Chen, Chen, Yu, Wan, and
  Wang}]{song-etal-2024-milebench}
Dingjie Song, Shunian Chen, Guiming~Hardy Chen, F.~Richard Yu, Xiang Wan, and
  Benyou Wang. 2024.
\newblock \href {https://arxiv.org/abs/2404.18532} {{MileBench}: Benchmarking
  {MLLMs} in long context}.
\newblock \emph{arXiv preprint arXiv:2404.18532}.

\bibitem[{Wang et~al.(2024{\natexlab{a}})Wang, Fu, Huang, Li, Liu, Liu, Ma, Xu,
  Zhou, Zhang et~al.}]{wang-etal-2024-muirbench}
Fei Wang, Xingyu Fu, James~Y. Huang, Zekun Li, Qin Liu, Xiaogeng Liu,
  Mingyu~Derek Ma, Nan Xu, Wenxuan Zhou, Kai Zhang, and 1 others.
  2024{\natexlab{a}}.
\newblock \href {https://arxiv.org/abs/2406.09411} {{MuirBench}: A
  comprehensive benchmark for robust multi-image understanding}.
\newblock \emph{arXiv preprint arXiv:2406.09411}.

\bibitem[{Wang et~al.(2024{\natexlab{b}})Wang, Shi, Tan, Qin, Wang, Zhang,
  Nambi, Ganu, and Wang}]{wang-etal-2024-mmneedle}
Hengyi Wang, Haizhou Shi, Shiwei Tan, Weiyi Qin, Wenyuan Wang, Tunyu Zhang,
  Akshay Nambi, Tanuja Ganu, and Hao Wang. 2024{\natexlab{b}}.
\newblock \href {https://arxiv.org/abs/2406.11230} {Multimodal needle in a
  haystack: Benchmarking long-context capability of multimodal large language
  models}.
\newblock \emph{arXiv preprint arXiv:2406.11230}.

\bibitem[{Wang et~al.(2024{\natexlab{c}})Wang, Zhang, Ren, Duan, Li, Liu, Hu,
  Chen et~al.}]{wang-etal-2024-mmniah}
Weiyun Wang, Shuibo Zhang, Yiming Ren, Yuchen Duan, Tiantong Li, Shuo Liu,
  Mengkang Hu, Zhe Chen, and 1 others. 2024{\natexlab{c}}.
\newblock \href {https://arxiv.org/abs/2406.07230} {Needle in a multimodal
  haystack}.
\newblock \emph{arXiv preprint arXiv:2406.07230}.

\bibitem[{Wang et~al.(2025)Wang, Yu, Ren, Zhang, Zhao, Saxena, Cheng, Wong,
  See, Minervini, Song, and Steedman}]{wang-etal-2025-mmlongbench}
Zhaowei Wang, Wenhao Yu, Xiyu Ren, Jipeng Zhang, Yu~Zhao, Rohit Saxena, Liang
  Cheng, Ginny Wong, Simon See, Pasquale Minervini, Yangqiu Song, and Mark
  Steedman. 2025.
\newblock \href {https://arxiv.org/abs/2505.10610} {{MMLongBench}: Benchmarking
  long-context vision-language models effectively and thoroughly}.
\newblock \emph{arXiv preprint arXiv:2505.10610}.

\bibitem[{Yang et~al.(2026)Yang, Shao, Huang, Dong, Liu, Tang, Zhou, Gao, Wang,
  Zhou, Yang, Wang, Sun, and Zhong}]{yang-etal-2026-courtsi}
Yuchen Yang, Yuqing Shao, Duxiu Huang, Linfeng Dong, Yifei Liu, Suixin Tang,
  Xiang Zhou, Yuanyuan Gao, Wei Wang, Yue Zhou, Xue Yang, Yanfeng Wang, Xiao
  Sun, and Zhihang Zhong. 2026.
\newblock \href {https://arxiv.org/abs/2603.09896} {Stepping {VLMs} onto the
  court: Benchmarking spatial intelligence in sports}.
\newblock \emph{arXiv preprint arXiv:2603.09896}.

\bibitem[{Yue et~al.(2024)Yue, Ni, Zhang, Zheng, Liu, Zhang, Stevens, Jiang,
  Ren, Sun et~al.}]{yue-etal-2024-mmmu}
Xiang Yue, Yuansheng Ni, Kai Zhang, Tianyu Zheng, Ruoqi Liu, Ge~Zhang, Samuel
  Stevens, Dongfu Jiang, Weiming Ren, Yuxuan Sun, and 1 others. 2024.
\newblock \href {https://arxiv.org/abs/2311.16502} {{MMMU}: A massive
  multi-discipline multimodal understanding and reasoning benchmark for expert
  {AGI}}.
\newblock In \emph{Proceedings of the IEEE/CVF Conference on Computer Vision
  and Pattern Recognition}.

\bibitem[{Zhang et~al.(2026)Zhang, Zhou, Li, Zhou, Shan, Lu, Cao, Chen, Han,
  Sheng, Tao, Liang, Wu, Shi, He, Lin, Zhang, Yan, Zhao, Li, Yu, Mei, Chen,
  Zhang, and Cui}]{zhang-etal-2026-browsecomp-v3}
Huanyao Zhang, Jiepeng Zhou, Bo~Li, Bowen Zhou, Yanzhe Shan, Haishan Lu,
  Zhiyong Cao, Jiaoyang Chen, Yuqian Han, Zinan Sheng, Zhengwei Tao, Hao Liang,
  Jialong Wu, Yang Shi, Yuanpeng He, Jiaye Lin, Qintong Zhang, Guochen Yan,
  Runhao Zhao, and 6 others. 2026.
\newblock \href {https://arxiv.org/abs/2602.12876} {{BrowseComp}-{$V^3$}: A
  visual, vertical, and verifiable benchmark for multimodal browsing agents}.
\newblock \emph{arXiv preprint arXiv:2602.12876}.

\bibitem[{Zhu et~al.(2021)Zhu, Lei, Huang, Wang, Zhang, Lv, Feng, and
  Chua}]{zhu-etal-2021-tatqa}
Fengbin Zhu, Wenqiang Lei, Youcheng Huang, Chao Wang, Shuo Zhang, Jiancheng Lv,
  Fuli Feng, and Tat-Seng Chua. 2021.
\newblock \href {https://arxiv.org/abs/2105.07624} {{TAT-QA}: A question
  answering benchmark on a hybrid of tabular and textual content in finance}.
\newblock In \emph{Proceedings of the 59th Annual Meeting of the Association
  for Computational Linguistics}.

\end{thebibliography}
	
	\appendix
	
	\section{Dataset-Level Details}
	\label{sec:dataset-details}
	
	Table~\ref{tab:token-buckets} reports the distribution of input token lengths for each subdataset.
	Token counts are computed with the Qwen3-VL-4B tokenizer over the normalized model input.
	The table shows that \benchmark{} covers both short visually grounded tasks and very long document-centered tasks, with financial reports, MMLongBench-Pic, PRISMM, and Pix2Fact contributing many examples above 16k tokens.

\begin{table*}[t]
	\centering
	\caption{Token-length buckets by subdataset, computed with the Qwen3-VL-4B tokenizer over normalized model inputs.}
	\label{tab:token-buckets}
	\resizebox{0.95\textwidth}{!}{%
	\renewcommand{\arraystretch}{1.10}
	\setlength{\tabcolsep}{6pt}
	\begin{tabular}{l | *{3}{r} | *{3}{r} | *{3}{r}}
		\toprule[1.2pt]
		\multirow{2}{*}{\textbf{Subdataset}} & \multicolumn{3}{c|}{\textbf{$\le$4k tokens}} & \multicolumn{3}{c|}{\textbf{4k--32k tokens}} & \multicolumn{3}{c}{\textbf{$>$32k tokens}} \\
		\cmidrule(lr){2-4} \cmidrule(lr){5-7} \cmidrule(lr){8-10}
		& \textbf{$\le$1k} & \textbf{1k--2k} & \textbf{2k--4k} & \textbf{4k--8k} & \textbf{8k--16k} & \textbf{16k--32k} & \textbf{32k--64k} & \textbf{64k--128k} & \textbf{$>$128k} \\
		\midrule[0.8pt]
		BrowseComp-V3 & 9 & 10 & 21 & 14 & 3 & 0 & 0 & 0 & 0 \\
		CL-Bench-Table & 6 & 5 & 62 & 11 & 9 & 2 & 1 & 0 & 0 \\
		CourtSI & 0 & 0 & 645 & 0 & 0 & 0 & 0 & 0 & 0 \\
		Financial Report ROE & 0 & 0 & 1 & 0 & 0 & 0 & 0 & 135 & 70 \\
		Insight-O3 & 1 & 13 & 28 & 15 & 9 & 79 & 0 & 0 & 0 \\
		MIRBench & 238 & 220 & 96 & 123 & 202 & 179 & 5 & 0 & 0 \\
		MMLongBench-Pic & 0 & 0 & 0 & 0 & 0 & 6 & 48 & 22 & 2 \\
		Paper Conclusion & 0 & 0 & 0 & 216 & 322 & 42 & 1 & 0 & 0 \\
		Pix2Fact & 0 & 0 & 0 & 0 & 7 & 36 & 0 & 0 & 0 \\
		PRISMM-Bench & 0 & 0 & 0 & 0 & 0 & 129 & 110 & 6 & 0 \\
		ReasonMap-Long & 0 & 0 & 0 & 17 & 2 & 82 & 0 & 0 & 0 \\
		ReasonMap-Plus & 0 & 0 & 0 & 14 & 2 & 21 & 0 & 0 & 0 \\
		ReasonMap-Short & 0 & 0 & 0 & 17 & 2 & 84 & 0 & 0 & 0 \\
		ZeroBench & 0 & 4 & 7 & 11 & 12 & 9 & 0 & 0 & 0 \\
		\bottomrule[1.2pt]
	\end{tabular}
	}
\end{table*}

\begin{figure*}[t]
	\centering
	\includegraphics[width=0.98\textwidth]{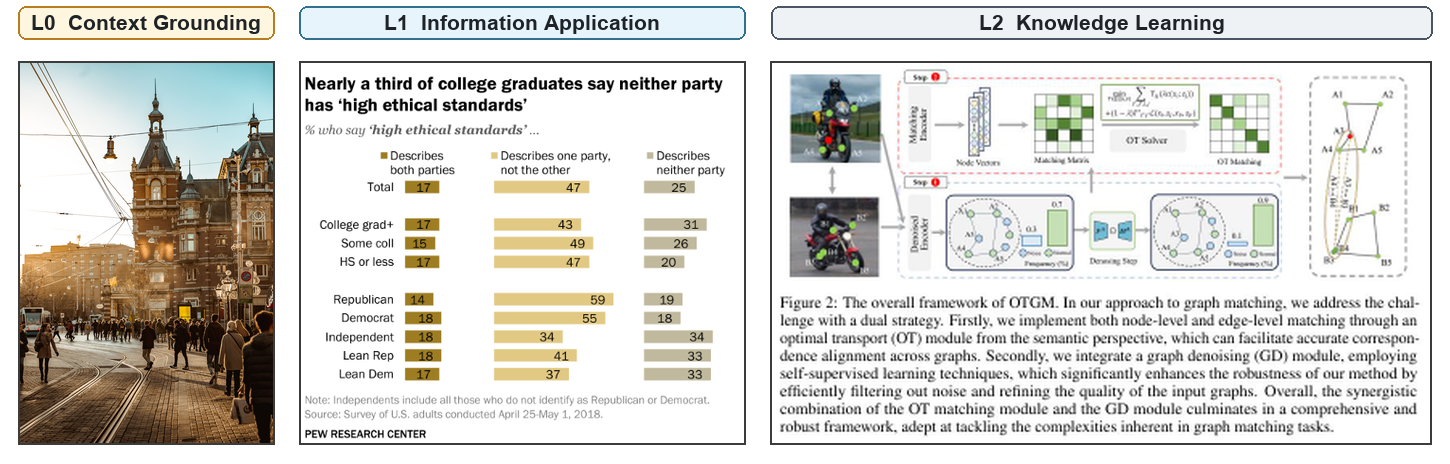}
	\caption{Appendix case-study panels aligned with the L0/L1/L2 taxonomy.
	Left (L0): visual grounding and entity-to-region binding error;
	middle (L1): context-specific row--column value binding error in chart reasoning;
	right (L2): failure to induce a context-defined procedural semantic in scientific-figure inconsistency repair.}
	\label{fig:appendix-case-study-triptych}
\end{figure*}

	\section{Prompt Templates}
	\label{sec:prompts}
	\begingroup
	\scriptsize
	
	We use a unified message-construction pipeline across all datasets.
	Each example provides a normalized user query and image list; runtime code optionally adds a system prompt, answer-format suffix, repair prompt, extraction prompt, or judge prompt.
	Image placeholders are written as \verb|<image>| and replaced by the corresponding image payload at inference time.
	Unless otherwise specified, the default system prompt is:
	\begin{verbatim}
You are a careful multimodal assistant.
Follow the user's output format exactly.
	\end{verbatim}
	
	\paragraph{Task prompt templates.}
	For public benchmark conversions, we preserve the original task query whenever possible and only add minimal answer-format constraints.
	The main templates are summarized below.
	\begin{verbatim}
Pix2Fact:
  Analyze the image(s), question, and crawled webpage
  information. Return a JSON object with Observation,
  Comprehensive Answer, and Final Answer.

MMLongBench-Pic:
  <document text and/or page images>
  <question>

PRISMM:
  You are provided with a full scientific paper that may
  contain an inconsistency.
  Question: What is the inconsistency in this paper?
  Options: A-D
  Please select the single best letter.

ReasonMap-Short / ReasonMap-Long:
  <image>
  According to the subway map, how do I get from
  <departure station> to <arrival station>?
  Return route segments with line name, departure stop,
  arrival stop, and via stops when required.

ReasonMap-Plus:
  <image>
  <counting, yes/no, numeric, or multiple-choice query>

InSight-O3:
  <image>
  <question>
  A. ... F. ...

MIRBench:
  Question: <question>
  Options: <options>
  Respond with:
  Option: <letter>
  Reason: <brief explanation>

ZeroBench:
  Question: <question>
  Put the final answer in curly braces: {final answer}.

CourtSI:
  <image>
  <sports-scene spatial question with required answer type>

BrowseComp-V3:
  <image(s)>
  [Question] <question>
  Reference Source [1..k]: <provided evidence>

CL-Bench:
  <task-specific system prompt>
  Component 2: Conversation Transcript
  <conversation transcript and supporting artifacts>
  <task request>
	\end{verbatim}
	
	For the two newly constructed datasets, we use the following templates.
	The financial-report prompt is originally written in Chinese; we show an English equivalent for readability.
	\begin{verbatim}
Financial Report ROE:
  Perform DuPont analysis for the provided financial report
  and compute ROE.
  Extract net profit, revenue, total assets, and equity.
  Compute profit margin, asset turnover, equity multiplier,
  and ROE. Show the calculation process.
  End with: ROE:<percentage with three significant digits>
  <financial report text and page images>

Paper Conclusion:
  You are an expert research analyst.
  Infer the main experimental conclusions of a research paper
  from the provided method description, experimental setup,
  result tables, and figures.
  Identify main comparisons and trends, then return 4-8 concise
  conclusions supported by the experiments.
	\end{verbatim}
	
	\paragraph{Answer-format and repair prompts.}
	For tasks with structured answers, we append a task-specific suffix such as:
	\begin{verbatim}
MCQ tasks: End with exactly one final line:
  Answer: <A/B/C/D/...>

Route tasks: Return only route segments:
  --
  Route Name: <line name>
  Departure Stop: <station>
  Arrival Stop: <station>
  Via Stops: <comma-separated stops, or None>

Numeric / short-answer tasks:
  End with exactly one final line:
  Answer: <short final answer>
	\end{verbatim}
	If a response cannot be parsed and format retry is enabled, the model receives:
	\begin{verbatim}
Your previous response could not be parsed for evaluation.
Task: <task>
Format issue: <reason>
Required format: <task-specific requirement>
Rewrite only the final answer in the required format.
Do not solve a different problem.
	\end{verbatim}
	
	\paragraph{Judge prompts.}
	For open-ended tasks, we use task-specific judges after extracting the final answer.
	Pix2Fact and general open-answer tasks use strict semantic equivalence prompts that output only \verb|True|/\verb|False| or \verb|<correct>|/\verb|<wrong>|.
	CL-Bench uses an all-or-nothing rubric judge: the answer receives 1 only if every rubric item is satisfied.
	Paper conclusion inference uses a list-count judge:
	\begin{verbatim}
Reference conclusions: <reference list>
Predicted conclusions: <prediction list>

Count how many predicted conclusions correctly match distinct
reference conclusions. Match by scientific claim, not wording.
Each reference and prediction can be matched at most once.
Output JSON:
{
  "correct_count": <integer>,
  "rationale": "<brief explanation>"
}
	\end{verbatim}
	
	\paragraph{Extraction prompts.}
	For free-form outputs, we first extract the final answer before scoring:
	\begin{verbatim}
Given a question and model response, extract only the final
answer. Prefer content inside <answer>...</answer> or boxed
expressions. Otherwise extract the last decisive answer span.
If there is no clear final answer, output [NO_FINAL_ANSWER].
	\end{verbatim}
	For MMLongBench, the extraction prompt additionally asks the extractor to return both the extracted answer and its type: integer, float, string, list, not answerable, or fail to answer.
	\endgroup

	\section{Additional Bad Cases}
	\label{sec:additional-badcases}
	
	\subsection{Failure Taxonomy Case Studies}
	\label{sec:failure-case-studies}
	
	To complement Section~6.2, we present three representative failures drawn from the case-study set in
	\texttt{write/case\_study}. Each case corresponds to one context-learning level and illustrates how an early mistake propagates into downstream answer errors.
	Figure~\ref{fig:appendix-case-study-triptych} summarizes the aligned visual evidence for the three cases (left: L0, middle: L1, right: L2).
	
	\paragraph{Case A (L0: Evidence Misbinding in Visual Grounding).}
	The left panel of Figure~\ref{fig:appendix-case-study-triptych} asks the model to find the board headed ``Zone Leidseplein,'' use its number $x$ to form $2011+x$, and retrieve China's Q1 year-on-year GDP growth for that year.
	The correct board reads ``Max. 14 dagen,'' so $x=14$, the year is 2025, and the answer is 5.4\%.
	The model instead grounds to a nearby white sign, extracts $x=102$, and returns no answer for 2113.
	Because its arithmetic and lookup are consistent after this initial mistake, the failure is L0 evidence misbinding rather than higher-level reasoning.
	
	\paragraph{Case B (L1: Context Misuse in Chart Value Binding).}
	In the middle panel, the model reaches the relevant Pew chart but misbinds the query slots for Democratic respondents and the ``neither party'' response.
	It returns a nearby value, 33\%, instead of the chart-supported 18\%.
	Retrieval and localization therefore succeed, while applying context-specific values to the correct target group and response category fails---an L1 context-misuse error.
	
	\paragraph{Case C (L2: Incomplete Induction of Context-Defined Procedure).}
	The right panel presents a scientific-paper inconsistency-repair task in which OT matching is followed by graph denoising.
	A correct answer requires inferring that changes in edge-noise composition must be reflected in the post-denoising graph topology.
	The model instead treats the problem as local layout checking and repositions the denoising module without updating the topology.
	This failure to induce and apply a paper-defined procedure is L2 incomplete induction, not a retrieval or slot-filling error.
	
	Together, the cases progress from incorrect evidence binding (L0), through incorrect value application (L1), to failure to learn and execute context-defined rules (L2), illustrating the diagnostic value of the three-level taxonomy.

\end{document}